\documentclass[preprint,preprintnumbers,amsmath,amssymb,superscriptaddress]{revtex4-2}

\usepackage{fullpage,amsthm,amsmath,amsfonts,bm,times,color,bbm}
\usepackage{hyperref}
\usepackage{graphicx}
\usepackage{mathtools}
\usepackage{caption}
\usepackage{booktabs}
\usepackage{multirow}
\usepackage{algorithmic}
\newcounter{algcounter}

\graphicspath{{figs/}}
\usepackage{soul}
\soulregister{\cite}{7} %
\soulregister{\ref}{7}  

\newcommand{\PMtwofive}{\ensuremath{\mathrm{PM}_{2.5}}}
\newcommand{\PMten}{\ensuremath{\mathrm{PM}_{10}}}
\newcommand{\um}{\ensuremath{\mu\mathrm{m}}}
\newcommand{\ugm}{\ensuremath{\mu\mathrm{g}/\mathrm{m}^{3}}}
\newcommand{\OmniPMNet}{OmniPM-Net}
\newcommand{\CAMS}{CAMS}
\newcommand{\GNN}{GNN}
\newcommand{\CTM}{CTM}
\newcommand{\ConvCNP}{ConvCNP}
\newcommand{\SSA}{SSA}
\newcommand{\AdaLN}{AdaLN}
\newcommand{\SetConv}{SetConv}
\newcommand{\SRTM}{SRTM}
\newcommand{\DEM}{DEM}
\newcommand{\GFS}{GFS}
\newcommand{\CNEMC}{CNEMC}
\newcommand{\OfficialUrbanStationCount}{1{,}618}
\newcommand{\StationArchiveCount}{1{,}618}
\newcommand{\MAE}{MAE}
\newcommand{\RMSE}{RMSE}
\newcommand{\POD}{POD}
\newcommand{\FAR}{FAR}
\newcommand{\CSI}{CSI}
\hypersetup{
  pdftitle={OmniPM-Net: Bridging discrete and gridded PM10 forecasts via omni-query neural processes},
  pdfsubject={PM10 air-quality forecast fusion via convolutional conditional neural processes},
}

\usepackage{easyReview}
\makeatletter\AtBeginDocument{\let\@elt\relax}\makeatother

\begin{document}

\title{
\OmniPMNet{}: Bridging discrete and gridded \texorpdfstring{\PMten{}}{PM10} forecasts via omni-query neural processes
}

\author{Shuangshuang He}
\affiliation{ColorfulClouds Technology Co., Ltd., 100875 Beijing, China.}
\email{heshuangshuang816@gmail.com}

\author{Shuo Wang}
\affiliation{School of Systems Science, Beijing Normal University, 100875 Beijing, China.}
\affiliation{D-ITET, ETH Zurich, 8092 Zurich, Switzerland.}

\begin{abstract}
	Forecasting particulate matter (\PMten{}) requires both station-scale accuracy and continuous spatial fields, especially during severe dust storms. Chemical transport models (\CTM{}s) provide gridded forecasts but retain local biases, whereas graph neural networks (\GNN{}s) track monitoring sites well at short lead times but do not produce gridded outputs. Here we present \OmniPMNet{}, a Convolutional Conditional Neural Process (\ConvCNP{})-based fusion model that reconciles these two forecast types within a shared spatial representation. A terrain-aware Gaussian set convolution lifts irregular \GNN{} station forecasts onto a regular grid, where a multi-scale Spatial Source Attention (\SSA{}) module blends them with Copernicus Atmosphere Monitoring Service (\CAMS{}) forecasts; a shared omni-query readout then decodes this representation into consistent \PMten{} predictions at either stations or grid cells over a 108\,h horizon. Evaluated across \StationArchiveCount{} air-quality monitoring stations throughout China over the full year of 2024, \OmniPMNet{} matches the station-level accuracy of the stronger \GNN{} baseline (mean absolute error 21.14 versus 22.00\,\ugm{}) and reduces the \CAMS{} mean absolute error by 30\%, while simultaneously delivering the gridded fields that the discrete \GNN{} cannot. Its clearest gains are in the high-concentration tail, where the 90th-percentile \MAE{} falls by 9\% relative to the \GNN{} and 25\% relative to \CAMS{}, and during dust episodes, where it improves categorical detection skill while tracking the evolving spatial trajectory of plumes and resolving local concentration peaks more closely than either input source. Source-attribution analysis reveals a physically meaningful, lead-time-dependent shift from \GNN{}-dominated short-range forecasts to greater \CAMS{} reliance at longer horizons. These results show that neural-process fusion can reconcile point and gridded forecasts within a common representation, offering a general approach for heterogeneous atmospheric data fusion.
\end{abstract}
\maketitle

\section{Introduction}
Particulate matter with an aerodynamic diameter below 10\,\um{} (\PMten{}) is a major contributor to air quality degradation and a key driver of regional climate forcing~\cite{anderson2012clearing,adar2014ambient}. \PMten{} originates from both anthropogenic emissions and natural sources such as windblown dust, and its concentration is shaped by highly nonlinear interactions among emission, transport, deposition, and complex topography, particularly during severe dust events driven by wind erosion over arid regions~\cite{ginoux2012global, shao2011dust, chen2017dust}.

Current approaches to \PMten{} forecasting face a spatial split: they either produce continuous gridded fields or discrete station-level predictions. Numerical chemical transport models (\CTM{}s)~\cite{seinfeld2016atmospheric}, such as the Copernicus Atmosphere Monitoring Service (\CAMS{})~\cite{inness2019cams} and WRF-Chem~\cite{grell2005wrfchem}, solve coupled atmospheric chemistry-transport equations to deliver regional forecasts on regular grids. However, uncertain emission inventories, coarse grid spacing, and parameterization errors leave \CTM{}s with systematic station-scale biases~\cite{flemming2015tropospheric,rao2020limit,gao2024review}. Although operational \CTM{} systems may assimilate observations, their forecast cycles and chemical initial-condition updates are not designed for rapid station-level correction during fast-evolving pollution events. Recent AI-driven models, from general medium-range weather systems~\cite{lam2023graphcast, bi2023pangu} to atmospheric-composition forecasters~\cite{bodnar2024aurora, gui2026advancing}, show strong global forecast skill but face a related limitation. Like traditional numerical weather prediction models, they are trained and evaluated on fixed grids. They therefore require interpolation to output predictions at irregular monitoring stations and can miss sub-grid variation governed by terrain and local emissions.

Conversely, graph neural network (\GNN{}) models~\cite{qi2019deep, wang2020pm25gnn, chen2021graph, wang2021attentive, hettige2024airphynet, li2023physics, pcdcnet2025} directly represent the spatial structure of monitoring stations. By conditioning predictions on the latest available station observations, \GNN{}s achieve high accuracy at the station level, particularly at short lead times. Yet, because \GNN{}s are fundamentally discrete, they produce forecasts only at predefined station locations and cannot generate the gridded spatial fields required for regional exposure mapping and plume tracking. 

Combining the spatial coverage of \CTM{}s with the local precision of \GNN{}s would help close this gap, but existing fusion methods usually compromise one source. Classical geostatistical approaches, such as Kriging with external drift or regression Kriging~\cite{cressie2015statistics,horalek2006spatial}, merge sparse station observations with gridded background fields through prescribed covariance models. These assumptions can over-smooth complex spatial heterogeneity. Deep-learning fusion approaches often force one modality into the representation space of the other. Discrete station measurements are pre-processed into continuous grids using inverse distance weighting (IDW) or ordinary Kriging before being passed to neural networks~\cite{zheng2015forecasting, le2020spatiotemporal}. Graph-based methods instead sample gridded \CTM{} fields only at station locations to construct node features~\cite{wang2020pm25gnn, hettige2024airphynet}. Statistical fusion methods combine heterogeneous estimates probabilistically, such as Bayesian ensembles that merge numerical-model output with satellite and statistical \PMtwofive{} estimates~\cite{murray2019bayesian}, while neural post-processing corrects numerical forecasts against observations~\cite{rasp2018neural}; both, however, are typically formulated for a single output representation rather than producing station and gridded forecasts together. These workarounds either introduce interpolation artifacts or discard gridded spatial continuity. Continuous fields and station networks therefore remain difficult to fuse in a way that preserves both local fidelity and regional coherence.

Neural processes (NPs)~\cite{garnelo2018conditional, kim2019attentive} offer a natural way to express this problem. They define prediction as a conditional mapping from irregular \textit{context} observations to arbitrary \textit{target} locations. Convolutional conditional neural processes (\ConvCNP{}s)~\cite{gordon2020convolutional} extend this idea by using translation-equivariant set convolutions to project irregular context points onto a regular functional space. Recent atmospheric applications point to the relevance of this formulation. \ConvCNP{}s have been used for local climate downscaling~\cite{vaughan2022convcnp}, simulation-to-reality frameworks for environmental NPs~\cite{scholz2023sim2real} align gridded numerical simulations with real sensor networks, and Aardvark~\cite{allen2025end} scales NP-style ideas to end-to-end weather prediction from sparse observations. These studies, however, ingest raw observations or single numerical fields and produce grid-only outputs. None addresses the joint encoding of two pre-existing forecast \emph{products} of different modalities---dynamic \GNN{} station forecasts and continuous \CAMS{} fields---or the simultaneous readout of point and gridded predictions from one shared representation.

Here, we present \textbf{\OmniPMNet{}}, an \textbf{omni-query} spatial fusion network that treats atmospheric data fusion as representation learning within the \ConvCNP{} framework (Fig.~\ref{fig:architecture}). \OmniPMNet{} maps irregular \GNN{} station forecasts into a gridded representation through a learnable, terrain-aware Gaussian set convolution, avoiding a fixed stationary covariance model. This representation is combined with \CAMS{} forecasts through a multi-scale Spatial Source Attention (\SSA{}) module with location-dependent weights. An omni-query readout then queries the shared latent field at arbitrary target coordinates, producing station-level and gridded \PMten{} forecasts from the same representation. Across the 2024 national urban monitoring network, including severe spring dust events, \OmniPMNet{} outperforms both input sources and shows an interpretable lead-time shift in source reliance, from station-informed \GNN{} forecasts at short ranges to \CAMS{} at longer ranges.

\section{Results}

\subsection*{Model overview}

\OmniPMNet{} fuses discrete \GNN{} station forecasts with \CAMS{} gridded forecasts into a shared latent representation on a regular grid. The model decodes this representation at arbitrary target coordinates, producing both station-level and continuous 0.25$^\circ$ gridded \PMten{} predictions over a 108\,h forecast horizon.
The framework builds on the Convolutional Conditional Neural Process (\ConvCNP{}), in which a model learns to map a set of \textit{context} observations at arbitrary locations to predictions at any \textit{target} location, adapted here to multi-source atmospheric forecast fusion.

\begin{figure}[h]
	\centering
	\includegraphics[width=\linewidth]{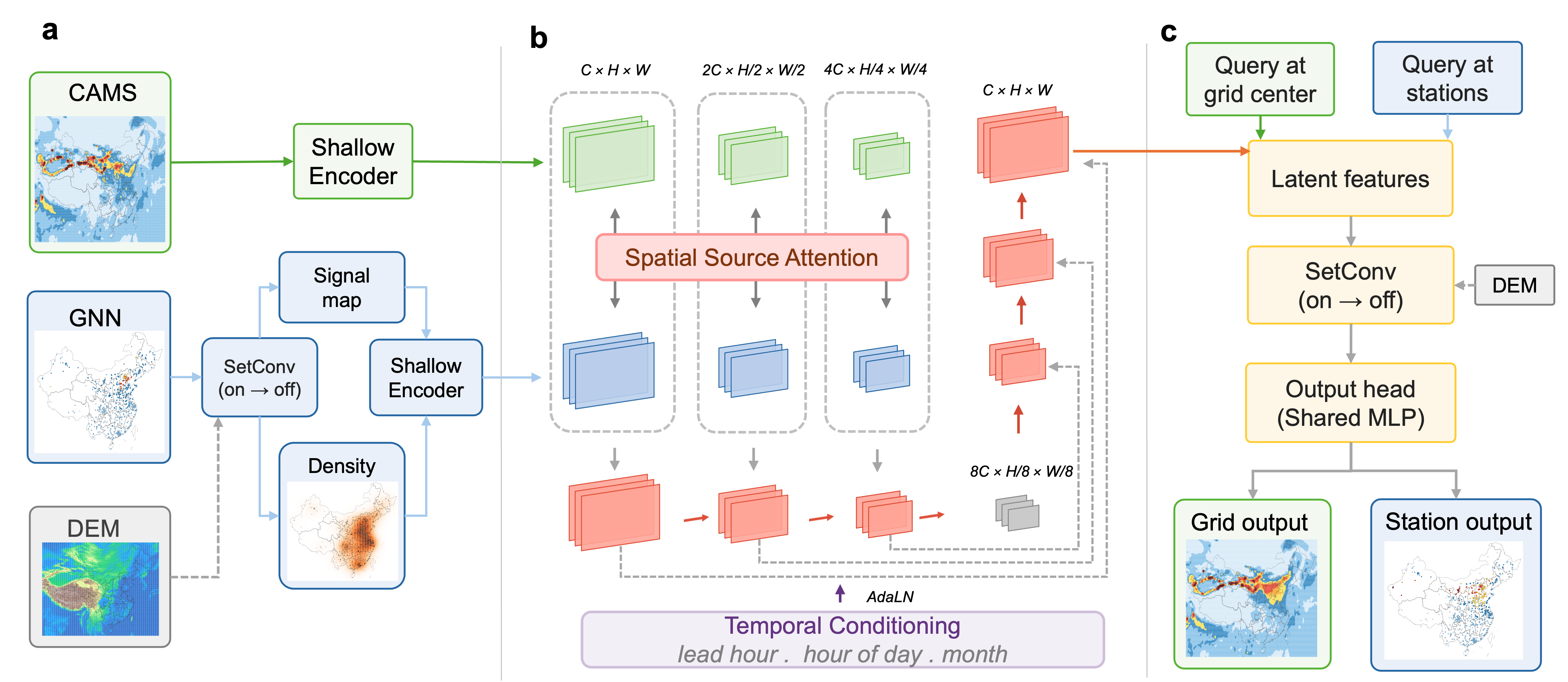}
	\captionsetup{labelformat=empty}
	\caption{}
	\label{fig:architecture}
\end{figure}

The architecture comprises three core modules (Fig.~\ref{fig:architecture}).

(1) \textbf{Per-source encoding alignment} (Fig.~\ref{fig:architecture}a). The discrete \GNN{} station forecasts, augmented with local \DEM{} elevation, are mapped onto a regular grid via a terrain-aware Gaussian set convolution with a learnable kernel bandwidth. This operation emits both a signal map and a density map, which are subsequently refined by a shallow convolutional encoder. In parallel, the continuous \CAMS{} gridded forecasts are processed by a separate shallow encoder, yielding two feature maps at the native grid resolution.

(2) \textbf{Dynamic multi-scale fusion} (Fig.~\ref{fig:architecture}b). The two encoded branches are progressively downsampled. At three distinct spatial scales, a Spatial Source Attention (\SSA{}) module dynamically blends them via multi-head cross-attention using location- and lead-time-dependent weights. Following a final bottleneck stage, the fused multi-scale features are decoded through a U-Net with skip connections. To capture temporal dynamics, all encoding and decoding layers are modulated by a temporal embedding (lead hour, hour of day, month) via adaptive layer normalization (\AdaLN{}).

(3) \textbf{Omni-query unified readout} (Fig.~\ref{fig:architecture}c). The unified latent feature map is queried at arbitrary target coordinates, accommodating both station locations and grid-cell centres, via a reverse set convolution. The extracted features are concatenated with the local \DEM{} elevation at the query point and passed through a shared MLP output head. Because both output paths traverse the same unified gridded representation and the same output head, spatial consistency between the two modalities is encouraged through joint representation and a consistency loss.

Each training sample randomly selects disjoint \textit{context} and \textit{target} subsets from the matched station panel. The \SetConv{} ingests only the context set, and the loss is computed exclusively on the target set. This masking strategy forces the network to learn a generalizable continuous spatial representation rather than memorizing the stationary geometry of the monitoring network. At inference time, all \StationArchiveCount{} station locations in the matched panel serve as context, while predictions can be queried at arbitrary target coordinates without retraining.

\subsection*{Overall forecast performance}

We evaluate \OmniPMNet{} against the single-source baselines (\GNN{} and \CAMS{}) over the full 2024 test set at \StationArchiveCount{} nationally controlled urban monitoring sites. As detailed in Table~\ref{tab:overall}, \OmniPMNet{} achieves the best score on every evaluated metric, although its aggregate margin over the stronger station-based \GNN{} baseline is modest and widens in the high-concentration tail. The overall mean absolute error (\MAE{}) drops to 21.14\,\ugm{}, a reduction of 4\% relative to the \GNN{} (22.00) and 30\% relative to \CAMS{} (30.02). The root-mean-square error (\RMSE{}) decreases to 58.44\,\ugm{}. The Pearson correlation rises from 0.48 for the \GNN{} and 0.33 for \CAMS{} to 0.55, indicating closer agreement with observations across the full year. \OmniPMNet{} also reduces the negative bias to $-$4.82\,\ugm{} and lowers the P90 \MAE{} from 80.11 (\GNN{}) and 97.47 (\CAMS{}) to 72.81\,\ugm{}. The largest relative gains therefore occur in the high-concentration tail, where both input forecasts tend to underestimate \PMten{}.

\begin{figure}[h]
	\centering
	\includegraphics[width=\linewidth]{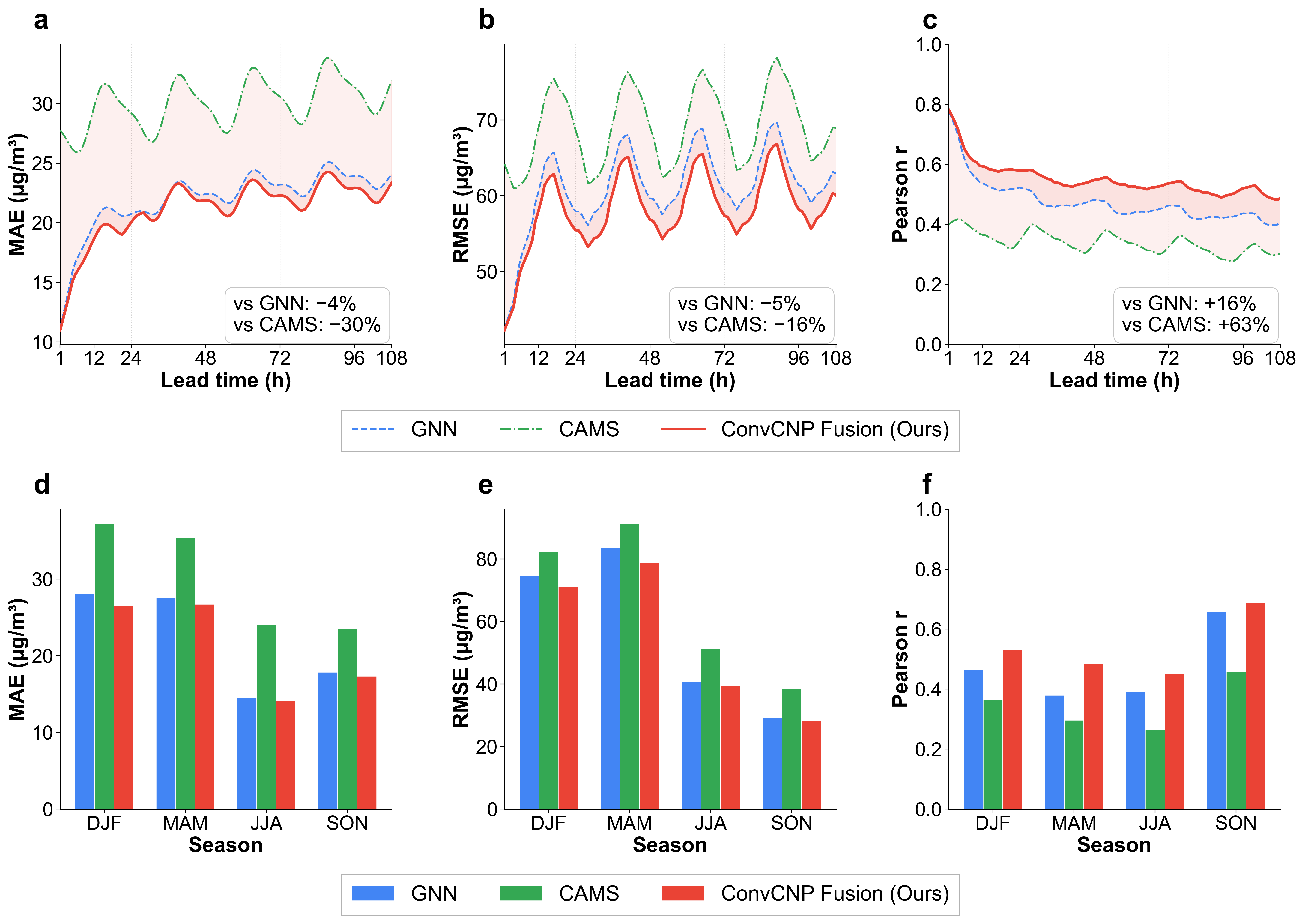}
	\captionsetup{labelformat=empty}
	\caption{}
	\label{fig:performance}
\end{figure}

Forecast accuracy changes with lead time (Fig.~\ref{fig:performance}a--c). During the initial 1--24\,h window, \OmniPMNet{} closely matches the \GNN{} baseline, preserving the short-range information carried by recent station observations. As the lead time extends to 25--72\,h, the \GNN{}'s accuracy degrades as the influence of initial observations weakens. \OmniPMNet{} instead uses the spatial continuity of \CAMS{} to keep a lower \MAE{} and higher Pearson correlation than either baseline. Beyond 72\,h, the \CAMS{} branch compensates for the fading \GNN{} skill and helps stabilize the forecast.

Seasonally, the highest absolute errors for all models occur in spring (MAM), when severe dust storms are frequent (Fig.~\ref{fig:performance}d--f). \OmniPMNet{} shows its largest relative improvement in this season, with the strongest \MAE{} and \RMSE{} reductions compared with \CAMS{}. In cleaner seasons such as summer (JJA), the model maintains similar performance to the stronger station-based baseline. These results indicate a consistent direction of improvement across regimes ranging from dust-dominated spring and winter periods to cleaner summer conditions.

\subsection*{Spatial error distribution}
\begin{figure}[h]
	\centering
	\includegraphics[width=\linewidth]{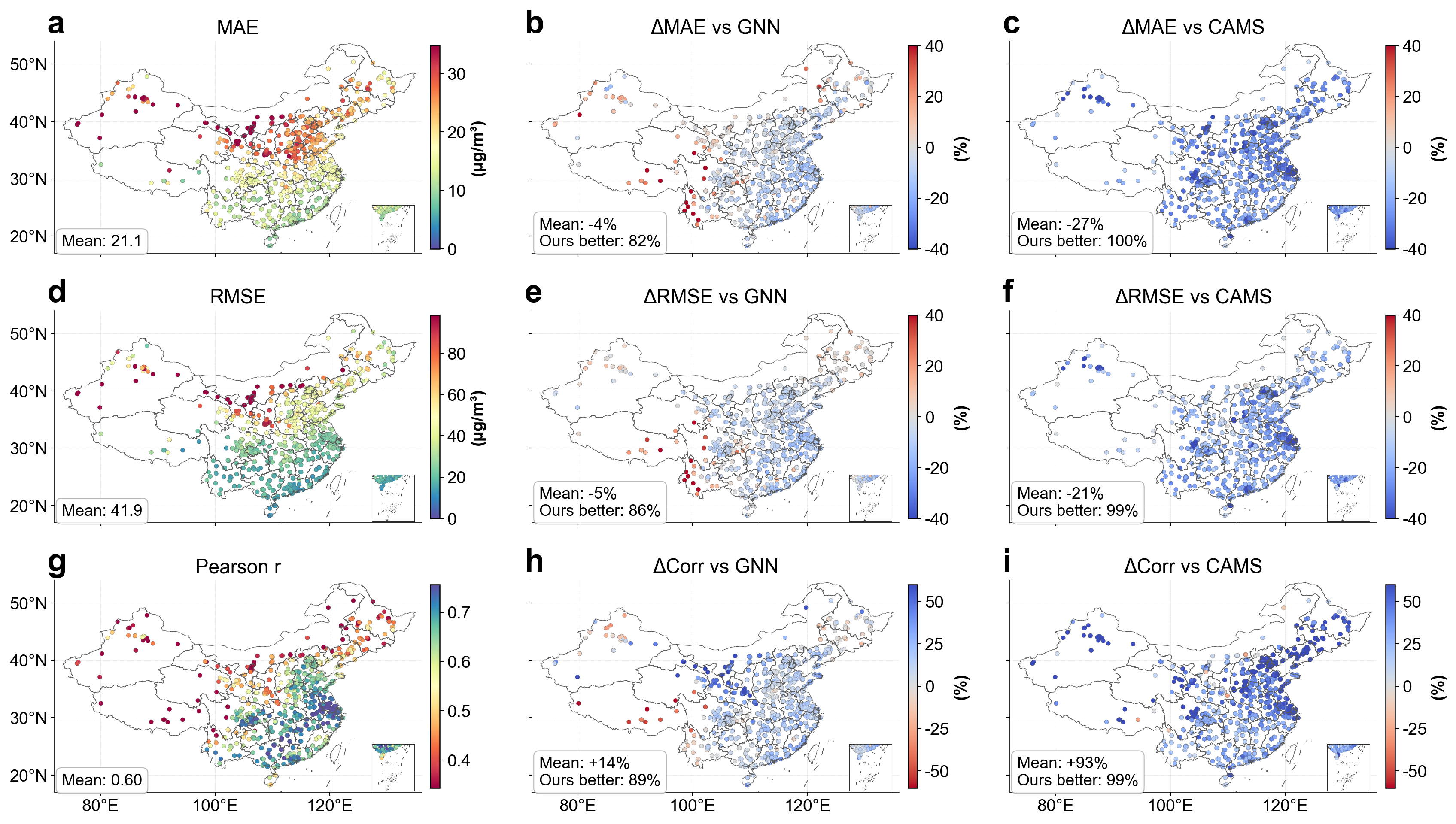}
	\captionsetup{labelformat=empty}
	\caption{}
	\label{fig:spatial}
\end{figure}

To examine the geographical structure of the error reductions, we map absolute forecast errors and their relative changes across China (Fig.~\ref{fig:spatial}). Figures~\ref{fig:spatial}a, d, and g display the absolute \MAE{}, \RMSE{}, and Pearson correlation of \OmniPMNet{}, while the subsequent panels present relative improvements compared to the baselines. 

As illustrated in Fig.~\ref{fig:spatial}a, the absolute \MAE{} exhibits a north-to-south gradient, with higher values in the dust-prone northwest and lower values along the southeastern coast. \OmniPMNet{} achieves a lower \MAE{} than the \GNN{} at 82\% of stations (Fig.~\ref{fig:spatial}b) and outperforms \CAMS{} at virtually all stations (Fig.~\ref{fig:spatial}c). The largest \MAE{} improvements relative to the \GNN{} are concentrated across central and eastern China, including the densely monitored Beijing--Tianjin--Hebei region and the broader North China Plain. In these areas, the fusion model uses the spatial continuity of \CAMS{} to enrich the \GNN{}'s discrete representations and correct coarse-resolution biases. In arid northwestern China and along the northern borders, performance locally degrades relative to the \GNN{}. These data-sparse peripheries provide limited nearby station support for the set convolution, and \CAMS{} background biases during near-source dust emissions can lower the fused station forecast. Even in these regions, however, \OmniPMNet{} remains well above \CAMS{} and provides spatial continuity where the discrete \GNN{} cannot produce gridded fields.

The \RMSE{} and Pearson correlation maps show similar spatial patterns (Fig.~\ref{fig:spatial}d--i). \OmniPMNet{} improves the \RMSE{} at 86\% of stations relative to the \GNN{} and at 99\% relative to \CAMS{}, yielding mean reductions of 5\% and 21\%, respectively. The Pearson correlation improves at 89\% and 99\% of stations, with mean increases of 14\% and 63\%. Because each station provides a paired comparison, these win rates correspond to two-sided binomial sign tests that reject parity with both baselines at $p < 0.001$ ($N = \StationArchiveCount{}$ stations) for all three metrics. The consistency of these gains across \MAE{}, \RMSE{}, and correlation indicates that the improvement is not an artifact of any single error statistic, and holds in data-sparse environments.

\subsection*{Severe dust-storm evaluation}
\begin{figure}[h]
	\centering
	\includegraphics[width=\linewidth]{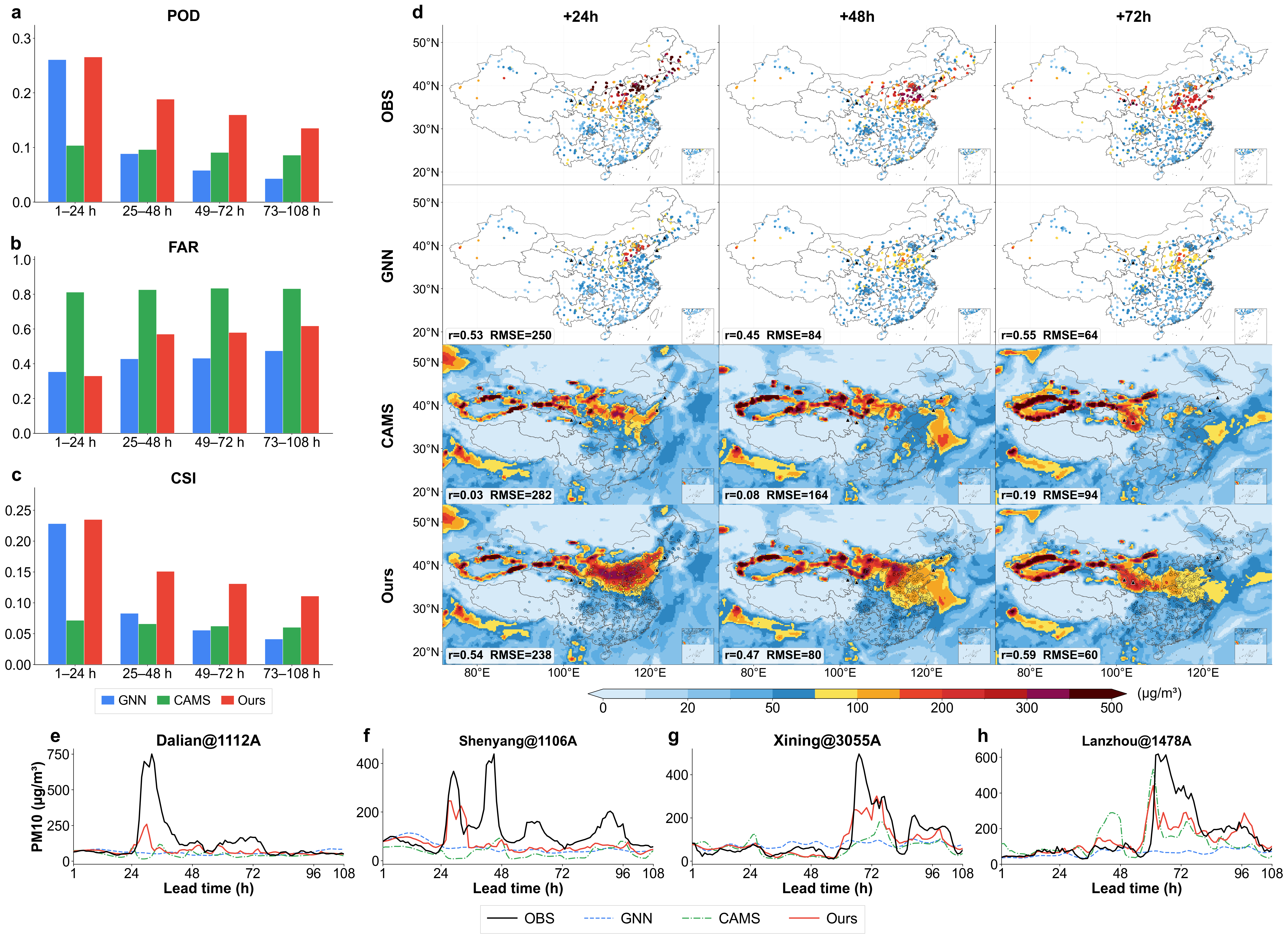}
	\captionsetup{labelformat=empty}
	\caption{}
	\label{fig:dust}
\end{figure}

Extreme dust events provide a stringent test of rapid, large-scale pollution transport. We conduct a categorical evaluation of severe high-\PMten{} episodes, defined by a fixed concentration threshold of 250\,\ugm{} applied throughout 2024; over northern China these extreme episodes are dominated by windblown dust, although the threshold itself does not distinguish dust from other sources. Figures~\ref{fig:dust}a--c show the probability of detection (\POD{}), false alarm ratio (\FAR{}), and critical success index (\CSI{}), respectively. The single-source baselines have opposite limitations. The discrete \GNN{} captures events at short range, but its \POD{} declines rapidly beyond 24\,h, leading to missed detections at longer lead times. \CAMS{} maintains a higher \POD{} but produces persistent false alarms throughout the forecast horizon (\FAR{} $>$ 0.8). \OmniPMNet{} reduces this trade-off by sustaining detection rates at longer lead times while suppressing many \CAMS{}-like false alarms. It therefore achieves the highest \CSI{} across all lead-time segments, with the clearest advantage beyond 24\,h.

We next examine the major Mongolian-origin dust storm of late March 2024 (27 March--2 April), which traversed northern China and produced extreme \PMten{} concentrations at many stations. Figure~\ref{fig:dust}d presents the spatial fields at three lead times (+24\,h, +48\,h, and +72\,h). 
The observations show substantial spatiotemporal evolution of the dust event over the 72-hour period. Initially spanning from Inner Mongolia across North and Northeast China (+24\,h), the high-\PMten{} core shifts toward the eastern coast by +48\,h, before reorganizing into dual centers over the Northwest and the North China Plain by +72\,h. The discrete \GNN{} substantially underestimates both the intensity and spatial extent of this outbreak, producing fragmented patterns with limited regional continuity. \CAMS{} captures the broad large-scale dust belt but exhibits noticeable spatial displacement of the high-concentration regions together with overly diffuse structures. In contrast, \OmniPMNet{} more accurately reproduces the evolving spatial trajectory and localized high-\PMten{} centers, yielding forecasts that are most consistent with the observations.

Time-series analysis at representative stations (Fig.~\ref{fig:dust}e--h) shows how the models differ across forecast horizons. At eastern sites such as Dalian and Shenyang (Fig.~\ref{fig:dust}e,f), rapid \PMten{} spikes occur early in the forecast window, approximately 24\,h in advance. Although \OmniPMNet{} slightly underestimates the absolute maxima, it anticipates the sharp onset and timing of the plume. Both single-source baselines miss these abrupt short-lead spikes. At western stations such as Xining and Lanzhou (Fig.~\ref{fig:dust}g,h), the extreme dust peaks occur later, at a lead time of roughly 60\,h. \CAMS{} captures the timing of these delayed events but has a pronounced negative bias. \OmniPMNet{} mitigates this underestimation and produces peak magnitudes closer to the observations.

\subsection*{Ablation studies and source attribution}

\begin{figure}[h]
	\centering
	\includegraphics[width=\linewidth]{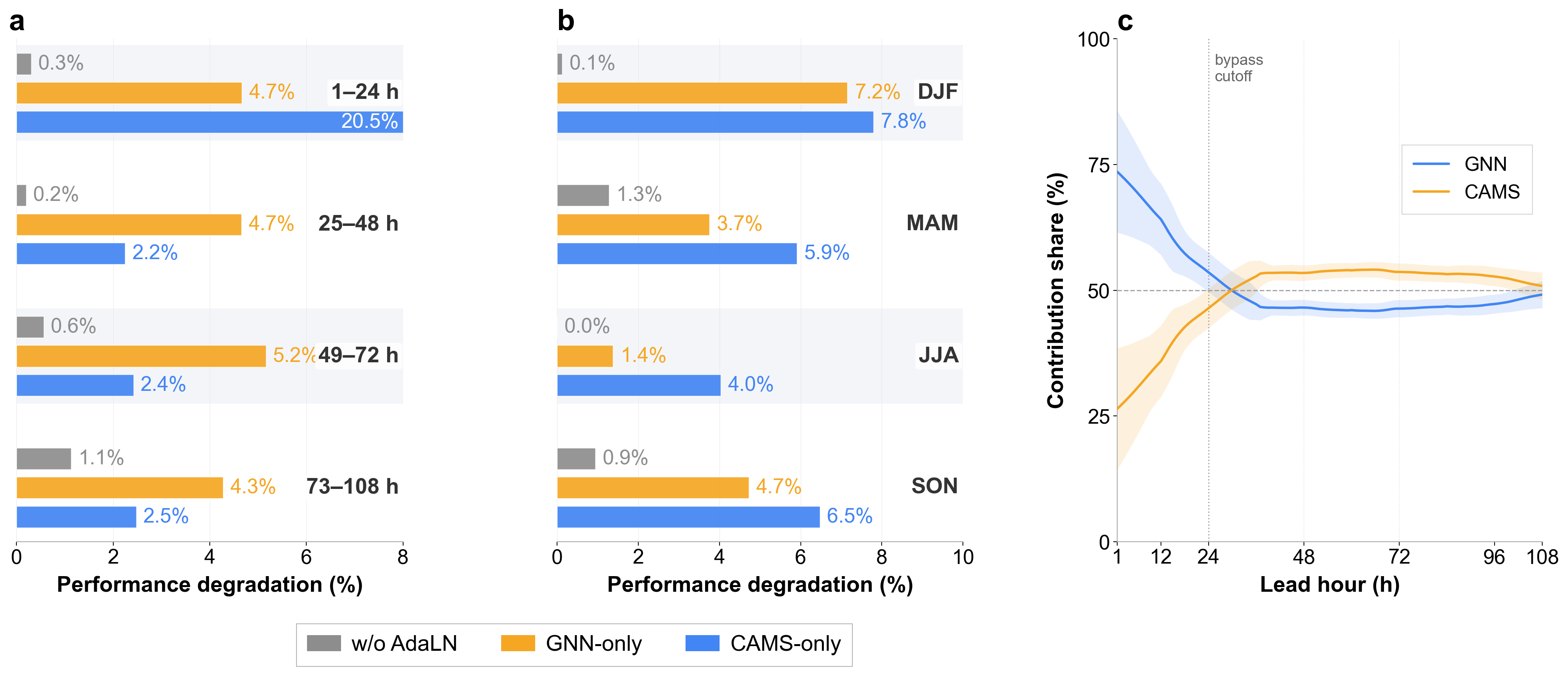}
	\captionsetup{labelformat=empty}
	\caption{}
	\label{fig:ablation}
\end{figure}

We quantify the predictive contribution of the two input data sources and the temporal conditioning module via ablation studies stratified by lead time (Fig.~\ref{fig:ablation}a) and season (Fig.~\ref{fig:ablation}b), together with a Shapley-based attribution analysis~\cite{shapley1953value,lundberg2017unified} that tracks each source's contribution across lead times (Fig.~\ref{fig:ablation}c).

Across the forecast horizon (Fig.~\ref{fig:ablation}a), ablating the temporal conditioning module (w/o \AdaLN{}) yields a consistent \MAE{} degradation of 0.2--1.1\%, peaking at the longest lead times (73--108\,h). Ablating the input data sources reveals a clear lead-time dependence. In the initial 1--24\,h window, removing the \GNN{} degrades the \MAE{} by 20.5\%, far exceeding the 4.7\% impact of removing \CAMS{}. Thus, short-range accuracy is mainly driven by the station-based \GNN{}. Beyond 24\,h, the influence of this observation-driven branch weakens: the degradation from removing the \GNN{} drops to 2.2--2.5\%, whereas the impact of removing \CAMS{} remains stable at 4.3--5.2\%. Consistent with the fading influence of the \GNN{}'s recent observational inputs at longer horizons, the continuous numerical background from \CAMS{} supplies more of the predictive information at extended lead times.

The impact of the temporal conditioning module also varies by season (Fig.~\ref{fig:ablation}b). Ablating the temporal conditioning module (w/o \AdaLN{}) yields a modest accuracy loss that peaks in spring (1.3\%) and autumn (0.9\%), but remains negligible in summer and winter. Because these temporal-conditioning effects (below 1.5\%) are comparable to the run-to-run variability expected of a single training seed, they should be read as suggestive only; by contrast, the large source-ablation effects (4--20\%) lie well outside this envelope. Meanwhile, removing the input data sources reveals distinct seasonal dependencies. In winter (DJF), removing either source causes a substantial accuracy loss (7.8\% for \GNN{}, 7.2\% for \CAMS{}). This indicates that both the station-based \GNN{} forecast and the gridded regional context from \CAMS{} contribute strongly to wintertime predictions, when \PMten{} levels and spatial transport are most intense. In summer (JJA), the asymmetry is pronounced (4.0\% degradation for removing \GNN{}, compared to 1.4\% for removing \CAMS{}), indicating that the fused summer forecast relies primarily on the station-based \GNN{}.

To quantify how each input data source contributes to the final prediction, we apply a Shapley-based attribution analysis~\cite{shapley1953value,lundberg2017unified} over the full 108\,h horizon (Fig.~\ref{fig:ablation}c). With only two input sources, the Shapley value reduces to the symmetrized marginal contribution: each source's share is derived from the change in target-station \MAE{} when that source is ablated by zeroing its input, normalized so that the two shares sum to 100\%. Because the \GNN{} branch is itself an observation-conditioned forecast that already ingests recent station measurements and \GFS{} meteorology, this attribution quantifies the contribution of each upstream \emph{product} rather than a separation of observations from atmospheric dynamics. The attribution curves mirror the lead-time dependence observed in the ablation studies. At +1\,h, the \GNN{} accounts for approximately 75\% of the attributed share, consistent with strong reliance on recent station observations. This share declines steeply over the first 24 hours, crossing parity with \CAMS{} near the 24\,h mark (vertical reference in Fig.~\ref{fig:ablation}c). In the medium-to-extended ranges, \CAMS{} maintains slightly more than half of the attribution share. The narrow $\pm 1\sigma$ envelopes, representing the spread across spatial locations and initialization dates rather than a seed-based confidence interval, indicate that this source shift is consistent across the test set.

\section{Discussion}
\OmniPMNet{} unifies sparse station forecasts and dense gridded fields through a shared latent representation on a regular grid. By lifting discrete station-level \GNN{} forecasts onto a continuous grid via a terrain-aware Gaussian set convolution, the model learns a spatial representation whose kernel bandwidth adapts to the monitoring network's layout. This learning-based mapping relaxes the fixed stationarity assumptions often invoked in classical geostatistics~\cite{cressie2015statistics}. Unlike a strictly translation-equivariant \ConvCNP{}, \OmniPMNet{} deliberately introduces location-dependent terrain conditioning, trading equivariance for the capacity to represent terrain- and location-specific biases. Its components---a Gaussian set convolution, multi-head cross-attention, adaptive layer normalization, and a U-Net backbone---are individually established; the contribution lies in their integration into a two-source neural process that emits both point and gridded forecasts. By fusing these representations through multi-scale Spatial Source Attention (\SSA{}) and decoding them through a shared omni-query readout, \OmniPMNet{} produces coherent station-level and gridded \PMten{} forecasts from a single forward pass. Compared with classical geostatistics and variational data assimilation, which rely on prescribed covariance structures or explicit dynamical operators, this approach learns the spatial prior directly from heterogeneous inputs.

The fusion framework behaves as a spatially and temporally adaptive map. In densely monitored regions, predictions remain closely tied to the observationally informed \GNN{} branch. In data-sparse interiors, the model relies more heavily on \CAMS{} for regional context. Local improvements over the \GNN{} can narrow in the most isolated settings because the set convolution receives limited nearby station support. For end users such as provincial environmental agencies, the gridded output supports exposure mapping and health advisories at locations without monitoring stations, and the detection skill sustained beyond 24\,h translates into multi-day lead for dust early warning, where \CAMS{}'s persistent false alarms (\FAR{} $>$ 0.8) and the \GNN{}'s short-range detection limit operational use. Even with this boundary, the architecture of independently encoding discrete and continuous inputs, fusing them at multiple scales, and decoding at arbitrary coordinates can be applied to other Earth system fusion tasks, such as merging radiosonde profiles with reanalysis fields or combining rain-gauge networks with radar-derived precipitation estimates.

The current formulation has several limitations. \OmniPMNet{} operates on two upstream forecast products rather than raw observations, so its accuracy is bounded by the quality of these inputs. If the \GNN{} severely underestimates an unprecedented event while \CAMS{} misrepresents its spatial structure, the fusion model has no independent information with which to correct the shared error. Our evaluation also benchmarks \OmniPMNet{} against its two input sources rather than against classical fusion methods such as Kriging or inverse-distance weighting, or a learned static blend; the specific contribution of the multi-scale \SSA{} and omni-query readout over simpler combination schemes therefore remains to be isolated. Quantitative evaluation of the gridded output is also confined to the national urban monitoring sites used for station-level verification. Because these sites are concentrated in urbanized areas, large rural and mountainous regions lack direct grid-scale validation. Pixel-level assessment against dense independent references, such as reanalysis fields or satellite aerosol optical depth retrievals, remains an important next step. The reported metrics also derive from a single training run; although the per-station improvements are statistically robust, a multi-seed quantification of run-to-run variability remains future work. Finally, the model parameters were optimized for the specific geometry of the monitoring network, and their generalizability to regions with different station densities or topographies requires further testing.

These limitations suggest a shift from \textit{forecast fusion} toward end-to-end \textit{observation-driven assimilation and forecasting}. The context-to-target formulation underlying \OmniPMNet{} can ingest raw multi-source observations, including ground-based air quality measurements, multi-sensor satellite aerosol retrievals, and numerical weather prediction fields. Treating these datasets as context sources at heterogeneous resolutions would allow them to be encoded within a single differentiable model and queried at any target coordinate. Such an extension could in principle exceed the accuracy bound imposed by relying on fixed upstream forecasts and would broaden the spatial coverage available for validation. It would also support multi-pollutant forecasting, linking assimilation-style context encoding with arbitrary-target prediction in one architecture.

\section*{Methods}

\subsection{Datasets}

The ground-truth labels for model training and evaluation are hourly \PMten{} concentrations from \CNEMC{}'s national urban air-quality real-time publishing platform~\cite{cnemcRealtimePlatform}. Official network documentation reports \OfficialUrbanStationCount{} nationally controlled urban air-quality monitoring sites across 339 cities, together with additional regional and background stations in the national network~\cite{mee2025monitoringNetwork}. We use this official urban-site count to construct a matched 2022--2024 panel of \StationArchiveCount{} monitoring sites by pairing observations, \GNN{} forecasts, and station metadata. The data are split chronologically by calendar year: observations from 2022 to 2023 form the development set, while the 2024 data are reserved exclusively for testing and metric evaluation. Early stopping and exponential-moving-average model selection use a chronological validation holdout drawn from the 2022--2023 development period, so the 2024 test year is never seen during training or model selection.

\OmniPMNet{} ingests two distinct sources of forecast data covering a forecast horizon of 1--108\,h. The discrete station-level inputs are produced by PCDCNet~\cite{pcdcnet2025}, a graph convolutional gated recurrent unit (GCGRU) model that acts as our \GNN{} baseline. It ingests 72\,h of historical air-quality observations at the national monitoring sites together with Global Forecast System (\GFS{}) meteorological guidance, providing hourly \PMten{} predictions at each station. The gridded inputs are taken from the Copernicus Atmosphere Monitoring Service (\CAMS{}) global atmospheric composition forecasts~\cite{inness2019cams}. While \CAMS{} provides twice-daily initializations (at 00\,UTC and 12\,UTC), we use the 12\,UTC forecast cycles, which extend to a 120\,h horizon. Because \CAMS{} products are released with an approximately 12\,h latency, only lead times beyond 12\,h would be accessible in real-time operational use. Nevertheless, we retain the full 1--108\,h range during training and evaluation to match the \GNN{} forecast horizon and enable direct method comparison; the aggregate metrics in Table~\ref{tab:overall} therefore include short leads (1--12\,h) that would not be operationally available from \CAMS{}.

To explicitly condition the spatial fusion on local terrain, we incorporate the Shuttle Radar Topography Mission (\SRTM{}) digital elevation model (\DEM{})~\cite{farr2007srtm}. For the gridded branch, the \DEM{} is aggregated to a 0.25$^\circ$ latitude--longitude model grid covering China (15$^\circ$--55$^\circ$\,N, 70$^\circ$--140$^\circ$\,E). Station elevations are sampled directly from the native-resolution \SRTM{} \DEM{} via bilinear interpolation at each station's coordinates, preserving sub-grid terrain detail at the monitoring sites.

\subsection{Data preprocessing}

The raw station observations undergo standard quality control to remove instrumental anomalies and physically implausible values. For the gridded inputs, the native 0.4$^\circ$ \CAMS{} fields are bilinearly interpolated to the 0.25$^\circ$ latitude--longitude model grid. Their mass-concentration units are simultaneously converted from kg/m$^{3}$ to \ugm{} to match the \GNN{} forecasts and station observations. To ensure stable gradient dynamics during training, all \PMten{} values, including \GNN{} forecasts, \CAMS{} fields, and ground-truth observations, are normalized by a common scaling factor of 200\,\ugm{}. This maps typical concentrations into a comparable dimensionless range. Similarly, the \SRTM{} \DEM{} is divided by a reference height of 6{,}000\,m to yield a dimensionless elevation field of comparable magnitude.

\subsection{Architecture details}


\OmniPMNet{} comprises three interconnected stages (Fig.~\ref{fig:architecture}). These include per-source encoding onto a common grid, multi-scale fusion and decoding, and an omni-query readout that produces both station-level and gridded predictions.

In the per-source encoding stage, \OmniPMNet{} employs a terrain-aware Gaussian \SetConv{} to bridge the spatial misalignment between discrete station coordinates and the model grid. Let $\mathcal{S}_{\mathrm{ctx}} = \{\mathbf{x}_i\}_{i=1}^{N_{\mathrm{ctx}}}$ denote the spatial coordinates of the context monitoring stations, a subset of the full station set $\mathcal{S}$. The input vector $\mathbf{y}_i = [\, c^{\mathrm{GNN}}_i, E_i \,]^\top$ concatenates the \GNN{} \PMten{} prediction $c^{\mathrm{GNN}}_i$ and the normalized \DEM{} elevation $E_i$. For any grid location $\mathbf{x}$, the \SetConv{} simultaneously produces a density map and an unnormalized signal, given by
\begin{equation}
	\rho(\mathbf{x}) = \sum_{i=1}^{N_{\mathrm{ctx}}} k(\mathbf{x} - \mathbf{x}_i; \sigma), \qquad
	\mathbf{s}(\mathbf{x}) = \sum_{i=1}^{N_{\mathrm{ctx}}} k(\mathbf{x} - \mathbf{x}_i; \sigma)\, \mathbf{y}_i,
	\label{eq:setconv}
\end{equation}
where $k(\mathbf{x} - \mathbf{x}_i; \sigma) = \exp(-\|\mathbf{x} - \mathbf{x}_i\|^2 / 2\sigma^2)$ is a Gaussian kernel with a learnable length scale $\sigma$, instantiated as $\sigma_{\mathrm{enc}}$ for this encoder set convolution (and as $\sigma_{\mathrm{dec}}$ for the decoder set convolution introduced below). The density $\rho(\mathbf{x})$ and signal $\mathbf{s}(\mathbf{x})$ are concatenated and passed through a shallow encoder, which learns the appropriate signal-to-density normalization rather than applying an explicit $\mathbf{s}/\rho$ division. This encoder consists of two $5\times5$ convolutions, where the second is dilated by a factor of two to expand the receptive field, followed by a $3\times3$ convolution. This process produces the \GNN{} feature map $F^{\mathrm{GNN}}$ on a $161 \times 281$ grid corresponding to a 0.25$^\circ$ resolution over China. In parallel, a separate shallow encoder comprising two $3\times3$ convolutions projects the single-channel \CAMS{} field into the matching feature space to produce $F^{\mathrm{CAMS}}$.

In the dynamic multi-scale fusion stage, each branch is independently downsampled through three U-Net~\cite{ronneberger2015unet} encoder stages with progressively increasing channel widths. This yields multi-scale feature maps $\{F^{\mathrm{GNN}}_\ell\}$ and $\{F^{\mathrm{CAMS}}_\ell\}$ for $\ell \in \{0, 1, 2\}$. At each of these three scales, a pixel-wise Spatial Source Attention (\SSA{}) mechanism dynamically blends the two feature maps. A depth-wise $3\times3$ convolution first enriches each branch with local spatial context. Subsequently, a shared query $Q$ is generated by projecting the average of the two branches concatenated with the lead-time embedding, expressed as
\begin{equation}
	Q = W_q\!\left( \left[\, \tfrac{1}{2}(F^{\mathrm{GNN}} + F^{\mathrm{CAMS}}); \tau_{\mathrm{emb}} \,\right] \right).
	\label{eq:query}
\end{equation}
The keys $(K_G, K_C)$ and values $(V_G, V_C)$ are linearly projected from $F^{\mathrm{GNN}}$ and $F^{\mathrm{CAMS}}$, respectively. Using multi-head scaled dot-product attention~\cite{vaswani2017attention} with four heads, the softmax-normalized attention weights $\alpha_G(\mathbf{x})$ and $\alpha_C(\mathbf{x})$, which sum to one across the two sources at each location, combine them with a residual connection to form the fused feature map
\begin{equation}
	F^{\mathrm{fused}} = W_{\mathrm{out}}\!\left(\alpha_G V_G + \alpha_C V_C\right) + \tfrac{1}{2}\left(F^{\mathrm{GNN}} + F^{\mathrm{CAMS}}\right).
	\label{eq:spatial_attn}
\end{equation}
To keep the model functional when either source is unreliable, we apply source dropout during training. With a probability of 0.05, one of the two branches is zeroed out for a given sample, forcing the model to remain usable under source failure. Following the multi-scale fusion, the deepest fused features $F^{\mathrm{fused}}_2$ are passed through a bottleneck module. This module performs a fourth spatial downsampling step followed by a residual block, maintaining a constant channel dimension. These deepest representations are processed by a symmetric three-stage U-Net decoder. The decoder uses skip connections exclusively from the fused feature maps $\{F^{\mathrm{fused}}_\ell\}_{\ell=0,1,2}$ rather than from the individual input branches. This yields the unified latent feature map $\mathbf{G} \in \mathbb{R}^{64 \times H \times W}$ on the $0.25^\circ$ model grid.

In the omni-query readout stage, \OmniPMNet{} produces predictions at any target coordinate $\mathbf{x}^\ast$ by querying $\mathbf{G}$ through a reverse \SetConv{} operation. This grid-to-point mapping uses its own learnable bandwidth $\sigma_{\mathrm{dec}}$. The sampled feature is concatenated with the \DEM{} elevation $E(\mathbf{x}^\ast)$ at the query point and projected through an elevation-merge linear layer with GELU activation. The augmented feature is then passed through a shared two-layer MLP head $h_\theta(\cdot)$ to obtain the model prediction $\hat{y}^{\mathrm{model}}(\mathbf{x}^\ast, \tau)$. Station-level predictions correspond to queries at the monitoring-site coordinates, whereas gridded predictions correspond to queries at all grid-cell centers. Because both outputs traverse the same latent feature map and output head, spatial consistency between the two modalities is encouraged through this joint representation and an auxiliary consistency loss.

To dynamically condition the network on the forecast horizon $\tau$, every convolutional layer in the encoder, bottleneck, and decoder is modulated by adaptive layer normalization (\AdaLN{}), defined as
\begin{equation}
	\mathrm{AdaLN}(h, \tau) = (\gamma_\tau + 1) \odot \mathrm{GroupNorm}(h) + \beta_\tau.
	\label{eq:adaln}
\end{equation}
The scale $\gamma_\tau$ and shift $\beta_\tau$ are produced by a two-layer MLP from a 128-dimensional temporal embedding $\tau_{\mathrm{emb}}$. This embedding concatenates sinusoidal encodings of the normalized lead time ($\tau / 108$), hour of day, and month.

\subsection{Training details}

\paragraph{Loss function.}
The model is trained using a composite loss function that balances station-level accuracy, gridded spatial structure, and dual-output consistency, defined as
\begin{equation}
	\mathcal{L} = \mathcal{L}_{\mathrm{Huber}}^{\mathrm{stn}} + \lambda_1\, \mathcal{L}_{\mathrm{SSIM}}^{\mathrm{grid}} + \lambda_2\, \mathcal{L}_{\mathrm{consist}}.
	\label{eq:loss}
\end{equation}
Here, $\mathcal{L}_{\mathrm{Huber}}^{\mathrm{stn}}$ is a Huber loss applied to the target-station predictions (with a transition point $\beta = 0.5$ in the 200\,\ugm{}-normalized units). This term incorporates a value-based weighting scheme that places stronger emphasis on high-concentration samples. Because high-resolution gridded ground truth is unavailable, $\mathcal{L}_{\mathrm{SSIM}}^{\mathrm{grid}}$ computes $1 - \mathrm{SSIM}$~\cite{wang2004ssim} between the model's gridded output and the \CAMS{} forecast over an $11 \times 11$ Gaussian window. This term acts as a soft physical regularizer, encouraging the model to retain the synoptic-scale plume morphology from \CAMS{} in data-sparse regions rather than generating spurious artifacts. A direct consequence is that, where station support is weak, the gridded output is anchored to \CAMS{} structure by construction; the off-station field therefore inherits \CAMS{} spatial priors and is not independently validated away from the monitoring sites (see Discussion). Finally, $\mathcal{L}_{\mathrm{consist}}$ is a Huber loss computed between the corresponding station observations and the gridded output bilinearly sampled at the target-station coordinates, distinct from the omni-query station readout supervised by $\mathcal{L}_{\mathrm{Huber}}^{\mathrm{stn}}$. This ties the two output modalities to a common reference at the supervised locations. We set the weighting coefficients to $\lambda_1 = 0.05$ and $\lambda_2 = 0.05$; these auxiliary weights were chosen heuristically to keep the regularizers subordinate to the primary station loss and were not extensively tuned.

\paragraph{Context and target sampling.}
Following the canonical neural-process training regime, each training sample randomly draws disjoint \textit{context} and \textit{target} sets from the matched station panel. Specifically, the context size is drawn uniformly from 800 to 1,200 stations, and the target size from 200 to 800 stations, subject to the stations available for that sample. The \SetConv{} ingests only the context set, and the loss is evaluated exclusively on the target set. This masking strategy forces the network to learn a generalizable continuous spatial representation rather than memorizing the stationary geometry of the monitoring network. During operational forecasting, all \StationArchiveCount{} station locations in the matched panel are ingested as the context set, and predictions can be queried at arbitrary target coordinates without retraining.

\paragraph{Implementation details.}
The network is optimized using the AdamW optimizer~\cite{loshchilov2017decoupled} with a batch size of 16, a weight decay of $10^{-5}$, and gradient clipping at a maximum norm of 1.0. The learning rate is initialized at $3 \times 10^{-4}$ and decayed to $10^{-6}$ following a cosine annealing schedule. To improve inference stability, an exponential moving average (EMA) of the model weights is maintained throughout training with a decay rate of 0.999. All final predictions and evaluations are produced using these EMA weights. The full model contains approximately 4.2 million trainable parameters and is trained for up to 30 epochs on a single NVIDIA A100 GPU, using an early stopping criterion with a patience of 15 epochs.

\subsection{Evaluation metrics}

All continuous metrics are computed at the monitoring station coordinates by comparing each method's point-level forecast against the corresponding hourly observation. Metrics pool the matched forecast--observation pairs over all evaluated initialization cycles of the 2024 test year; station-hours lacking a valid observation or a valid \GNN{} or \CAMS{} value are excluded from all methods identically. We report the mean absolute error (\MAE{}), root-mean-square error (\RMSE{}), systematic bias (defined as the mean prediction minus the mean observation), 90th-percentile \MAE{} (P90 \MAE{}, the \MAE{} computed over the decile of station-hours with the highest observed \PMten{}, pooled across all stations and lead times), and the Pearson correlation coefficient ($r$).

To assess the model's capability in capturing extreme pollution episodes, we conduct a categorical evaluation using a fixed \PMten{} threshold of 250\,\ugm{}. This threshold flags severe high-\PMten{} exceedances, which over northern China in spring are predominantly dust-driven; because the evaluation uses no aerosol speciation, it does not formally isolate dust from other extreme-\PMten{} sources. A hit, miss, or false alarm is scored per station-hour by thresholding both the forecast and the observation at 250\,\ugm{}, and the probability of detection (\POD{}), false alarm ratio (\FAR{}), and critical success index (\CSI{}) are computed from the pooled counts within each lead-time segment (1--24, 25--48, 49--72, and 73--108\,h).

Finally, seasonal performance is decomposed according to standard meteorological quarters: winter (DJF; December, January, and February), spring (MAM; March, April, and May), summer (JJA; June, July, and August), and autumn (SON; September, October, and November).

\section*{Data availability}

The datasets analyzed during the current study are derived from public repositories. \CAMS{} global atmospheric composition forecasts are publicly available through the Copernicus Atmosphere Data Store (\url{https://ads.atmosphere.copernicus.eu/}). Hourly \PMten{} observations from China's national monitoring network are accessible via the China National Environmental Monitoring Centre (\CNEMC{}) platform (\url{http://www.cnemc.cn/}). Global Forecast System (\GFS{}) meteorological data are provided by the NOAA National Centers for Environmental Information (\url{https://www.ncei.noaa.gov/}). \SRTM{} elevation data can be accessed through the USGS EarthExplorer (\url{https://earthexplorer.usgs.gov/}). 

\section*{Code availability}

The source code for the \OmniPMNet{} model and evaluation scripts will be made publicly available on GitHub upon acceptance.

\section*{Acknowledgments}

[To be added.]

\section*{Author contributions}

S.H. conceived the study, curated the datasets, developed and trained the models, performed the evaluations, produced the figures, and drafted the manuscript.
S.W. contributed to dataset curation and manuscript revision.
Both authors discussed the results and approved the final version of the manuscript.



\section*{Competing interests}

The authors declare no competing interests.

\bibliographystyle{naturemag}

\clearpage
\begin{table}[h]
	\captionsetup{labelformat=empty}
	\caption{}
	\centering
	\begin{tabular}{lccccc}
		\toprule
		Method & \MAE{} & \RMSE{} & Bias & P90 \MAE{} & Pearson $r$ \\
			& (\ugm{}) & (\ugm{}) & (\ugm{}) & (\ugm{}) & \\
		\midrule
		\GNN{}              & \underline{22.00} & \underline{61.39} & \underline{$-$5.59} & \underline{80.11} & \underline{0.479} \\
		\CAMS{}             & 30.02 & 69.32 & $-$8.32 & 97.47 & 0.334 \\
		\textbf{\OmniPMNet{}} & \textbf{21.14} & \textbf{58.44} & \textbf{$-$4.82} & \textbf{72.81} & \textbf{0.549} \\
		\bottomrule
	\end{tabular}
	\label{tab:overall}
\end{table}
\noindent \textbf{TABLE~I. Overall forecast performance (all lead times, 1--108\,h).}
\MAE{}, \RMSE{}, systematic bias, 90th-percentile \MAE{} (P90 \MAE{}), and Pearson correlation coefficient $r$ averaged over \StationArchiveCount{} station locations and the full 2024 test year. Best values are shown in \textbf{bold}; second-best are \underline{underlined}.

\clearpage
\noindent \textbf{FIG.~1. Architecture of \OmniPMNet{}.}
\textbf{(a) Per-source encoding alignment.} The discrete \GNN{} station forecasts are mapped onto a regular grid by a terrain-aware Gaussian \SetConv{} that emits both a signal map and a density map. These are subsequently refined by a shallow encoder. In parallel, the \CAMS{} gridded forecasts are encoded by a separate shallow encoder.
\textbf{(b) Multi-scale fusion and decoding.} The two encoded branches are progressively downsampled. At three distinct spatial scales, a Spatial Source Attention (\SSA{}) module blends them via multi-head cross-attention using location- and lead-time-dependent weights. Following a final bottleneck stage, the fused features are decoded through a U-Net with skip connections. All encoder and decoder layers are modulated by a temporal embedding (lead hour, hour of day, month) through adaptive layer normalization (\AdaLN{}).
\textbf{(c) Omni-query readout.} The shared latent feature map is queried at arbitrary target coordinates, accommodating both grid centers and station locations, via a reverse \SetConv{}. The extracted features are concatenated with the local \DEM{} elevation at the query point and passed through a shared MLP output head to produce gridded and station-level forecasts simultaneously.

\clearpage
\noindent \textbf{FIG.~2. Overall forecast performance.}
\textbf{a--c,} Mean absolute error (\MAE{}), root-mean-square error (\RMSE{}), and Pearson correlation coefficient ($r$) as a function of lead time for the \GNN{} (blue dashed), \CAMS{} (green dashed), and \OmniPMNet{} (red solid). Annotated percentages denote the relative improvement of the fusion model over the single-source baselines.
\textbf{d--f,} Seasonal variations of \MAE{}, \RMSE{}, and Pearson $r$ (DJF, MAM, JJA, SON) for the three methods. \OmniPMNet{} achieves its most pronounced relative improvement during the dust-dominated spring season (MAM). All metrics are evaluated across \StationArchiveCount{} station locations using the complete 2024 test set.

\clearpage
\noindent \textbf{FIG.~3. Spatial distribution of station-level forecast performance.}
\textbf{a, d, g,} Absolute \MAE{}, \RMSE{}, and Pearson $r$ of \OmniPMNet{} at each national urban monitoring site.
\textbf{b, e, h,} Relative changes in \MAE{}, \RMSE{}, and Pearson $r$ compared to the \GNN{} (blue indicates \OmniPMNet{} advantage; red indicates \GNN{} advantage). \OmniPMNet{} outperforms the \GNN{} at 82\%, 86\%, and 89\% of stations for \MAE{}, \RMSE{}, and $r$, respectively.
\textbf{c, f, i,} Relative changes compared to \CAMS{}. \OmniPMNet{} outperforms \CAMS{} at $\geq$99\% of stations across all three metrics.

\clearpage
\noindent \textbf{FIG.~4. Dust-storm forecast skill and a case study (27 March--2 April 2024).}
\textbf{a--c,} Probability of detection (\POD{}), false alarm ratio (\FAR{}), and critical success index (\CSI{}) for severe high-\PMten{} episodes (threshold 250\,\ugm{}) over 2024, stratified by lead-time segment (1--24, 25--48, 49--72, and 73--108\,h).
\textbf{d,} Spatial fields at +24\,h, +48\,h, and +72\,h lead times for ground-truth observations (top row), the \GNN{} (second row), \CAMS{} (third row), and \OmniPMNet{} (bottom row). Regions without observations are zero-filled for visualization only; the annotated station-scale \RMSE{} and Pearson $r$ are single-initialization case-study values computed only at observed stations and exclude these zero-filled cells.
\textbf{e--h,} Time series at representative stations along the dust-transport pathway: Dalian (\textbf{e}), Shenyang (\textbf{f}), Xining (\textbf{g}), and Lanzhou (\textbf{h}). Black lines indicate observations, blue the \GNN{}, green \CAMS{}, and red \OmniPMNet{}.

\clearpage
\noindent \textbf{FIG.~5. Ablation studies and Shapley-based source attribution.}
\textbf{a,} Performance degradation (\%) of ablated configurations relative to the full model, stratified by lead time (1--24, 25--48, 49--72, and 73--108\,h). The evaluated configurations include removing the \GNN{} input (\CAMS{}-only, blue), removing the \CAMS{} input (\GNN{}-only, orange), and ablating the temporal conditioning module (w/o \AdaLN{}, grey).
\textbf{b,} Performance degradation of the ablated configurations stratified by season (DJF, MAM, JJA, SON).
\textbf{c,} Shapley-based contribution share (\%) of the \GNN{} (blue) and \CAMS{} (orange) inputs across the 108\,h forecast horizon. Shaded envelopes represent $\pm 1\sigma$ variations across locations and initialization dates, and the vertical dotted line denotes the 24\,h mark.

\clearpage

\bibliography{references}

@book{seinfeld2016atmospheric,
  title     = {Atmospheric Chemistry and Physics: From Air Pollution to Climate Change},
  author    = {Seinfeld, John H. and Pandis, Spyros N.},
  publisher = {John Wiley \& Sons},
  edition   = {3rd},
  year      = {2016},
}

@article{rao2020limit,
  title   = {On the limit to the accuracy of regional-scale air quality models},
  author  = {Rao, S. Trivikrama and Luo, Huiying and Astitha, Marina and Hogrefe, Christian and Garcia, Valerie and Mathur, Rohit},
  journal = {Atmospheric Chemistry and Physics},
  volume  = {20},
  number  = {3},
  pages   = {1627--1639},
  year    = {2020},
  doi     = {10.5194/acp-20-1627-2020},
}

@article{gao2024review,
  title   = {A review of the {CAMx}, {CMAQ}, {WRF-Chem} and {NAQPMS} models: Application, evaluation and uncertainty factors},
  author  = {Gao, Zhaoqi and Zhou, Xuehua},
  journal = {Environmental Pollution},
  volume  = {343},
  pages   = {123183},
  year    = {2024},
  doi     = {10.1016/j.envpol.2023.123183},
}

@article{grell2005wrfchem,
  title   = {Fully coupled ``online'' chemistry within the {WRF} model},
  author  = {Grell, Georg A. and Peckham, Steven E. and Schmitz, Rainer and McKeen, Stuart A. and Frost, Gregory and Skamarock, William C. and Eder, Brian},
  journal = {Atmospheric Environment},
  volume  = {39},
  number  = {37},
  pages   = {6957--6975},
  year    = {2005},
  doi     = {10.1016/j.atmosenv.2005.04.027},
}

@article{anderson2012clearing,
  title   = {Clearing the air: a review of the effects of particulate matter air pollution on human health},
  author  = {Anderson, J. O. and Thundiyil, J. G. and Stolbach, A.},
  journal = {Journal of Medical Toxicology},
  volume  = {8},
  number  = {2},
  pages   = {166--175},
  year    = {2012},
  doi     = {10.1007/s13181-011-0203-1},
}

@article{adar2014ambient,
  title   = {Ambient coarse particulate matter and human health: a systematic review and meta-analysis},
  author  = {Adar, Sara D. and Filigrana, Paola A. and Clements, Nicholas and Peel, Jennifer L.},
  journal = {Current Environmental Health Reports},
  volume  = {1},
  number  = {3},
  pages   = {258--274},
  year    = {2014},
  doi     = {10.1007/s40572-014-0022-z},
}

@article{ginoux2012global,
  title   = {Global-scale attribution of anthropogenic and natural dust sources and their emission rates based on {MODIS Deep Blue} aerosol products},
  author  = {Ginoux, Paul and Prospero, Joseph M. and Gill, Thomas E. and Hsu, N. Christina and Zhao, Ming},
  journal = {Reviews of Geophysics},
  volume  = {50},
  number  = {3},
  pages   = {RG3005},
  year    = {2012},
  doi     = {10.1029/2012RG000388},
}

@article{shao2011dust,
  title   = {Dust cycle: An emerging core theme in {Earth} system science},
  author  = {Shao, Yaping and Wyrwoll, Karl-Heinz and Chappell, Adrian and Huang, Jianping and Lin, Zhaohui and McTainsh, Grant H. and Mikami, Masao and Tanaka, Taichu Y. and Wang, Xulong and Yoon, Soonchang},
  journal = {Aeolian Research},
  volume  = {2},
  number  = {4},
  pages   = {181--204},
  year    = {2011},
  doi     = {10.1016/j.aeolia.2011.02.001},
}

@article{chen2017dust,
  title   = {Emission, transport, and radiative effects of mineral dust from the {Taklimakan} and {Gobi} deserts: comparison of measurements and model results},
  author  = {Chen, Siyu and Huang, Jianping and Kang, Litai and Wang, Hao and Ma, Xiaojun and He, Yongli and Yuan, Tiangang and Yang, Ben and Huang, Zhongwei and Zhang, Guolong},
  journal = {Atmospheric Chemistry and Physics},
  volume  = {17},
  number  = {3},
  pages   = {2401--2421},
  year    = {2017},
  doi     = {10.5194/acp-17-2401-2017},
}

@article{inness2019cams,
  title   = {The {CAMS} reanalysis of atmospheric composition},
  author  = {Inness, Antje and Ades, Melanie and Agust{\'\i}-Panareda, Anna and Barr{\'e}, J{\'e}r{\^o}me and Benedictow, Anna and Blechschmidt, Anne-Marlene and Dominguez, Juan Jose and Engelen, Richard and Eskes, Henk and Flemming, Johannes and others},
  journal = {Atmospheric Chemistry and Physics},
  volume  = {19},
  number  = {6},
  pages   = {3515--3556},
  year    = {2019},
  doi     = {10.5194/acp-19-3515-2019},
}

@misc{cnemcRealtimePlatform,
  title  = {National Urban Air Quality Real-Time Publishing Platform},
  author = {{China National Environmental Monitoring Centre}},
  year   = {2026},
  url    = {https://air.cnemc.cn:18007/},
  note   = {Accessed 9 June 2026},
}

@techreport{mee2025monitoringNetwork,
  title       = {Compilation Notes for the Draft Technical Specification on Data Acquisition and Transmission for Ambient Air Quality Automatic Monitoring Systems},
  author      = {{Ministry of Ecology and Environment of the People's Republic of China}},
  institution = {Ministry of Ecology and Environment of the People's Republic of China},
  year        = {2025},
  month       = {August},
  url         = {https://www.mee.gov.cn/xxgk2018/xxgk/xxgk06/202509/W020250905372030791280.pdf},
  note        = {In Chinese},
}

@article{flemming2015tropospheric,
  title   = {Tropospheric chemistry in the {Integrated Forecasting System} of {ECMWF}},
  author  = {Flemming, Johannes and Huijnen, Vincent and Arteta, Joaquim and Bechtold, Peter and Beljaars, Anton and Blechschmidt, A-M and Diamantakis, Michail and Engelen, Richard J and Gaudel, Audrey and Inness, Antje and others},
  journal = {Geoscientific Model Development},
  volume  = {8},
  number  = {4},
  pages   = {975--1003},
  year    = {2015},
  doi     = {10.5194/gmd-8-975-2015},
}

@article{bodnar2024aurora,
  title   = {A foundation model for the {Earth} system},
  author  = {Bodnar, Cristian and Bruinsma, Wessel P. and Lucic, Ana and Stanley, Megan and Brandstetter, Johannes and Garvan, Patrick and Riechert, Maik and Weyn, Jonathan and Dong, Haiyu and Vaughan, Anna and others},
  journal = {Nature},
  volume  = {641},
  number  = {8065},
  pages   = {1180--1187},
  year    = {2025},
  doi     = {10.1038/s41586-025-09005-y},
}

@article{gui2026advancing,
  title   = {Advancing operational global aerosol forecasting with machine learning},
  author  = {Gui, Ke and Zhang, Xutao and Che, Huizheng and Li, Lei and Zheng, Yu and An, Linchang and Miao, Yucong and Zhao, Hujia and Dubovik, Oleg and Holben, Brent and others},
  journal = {Nature},
  volume  = {651},
  pages   = {658--665},
  year    = {2026},
  doi     = {10.1038/s41586-026-10234-y},
}

@article{qi2019deep,
  title   = {A hybrid model for spatiotemporal forecasting of {PM}$_{2.5}$ based on graph convolutional neural network and long short-term memory},
  author  = {Qi, Yanlin and Li, Qi and Karimian, Hamed and Liu, Di},
  journal = {Science of the Total Environment},
  volume  = {664},
  pages   = {1--10},
  year    = {2019},
  doi     = {10.1016/j.scitotenv.2019.01.333},
}

@article{chen2021graph,
  title   = {{Group-Aware Graph Neural Network} for nationwide city air quality forecasting},
  author  = {Chen, Ling and Xu, Jiahui and Wu, Binqing and Huang, Jianlong},
  journal = {ACM Transactions on Knowledge Discovery from Data},
  volume  = {18},
  number  = {3},
  pages   = {1--20},
  year    = {2023},
  doi     = {10.1145/3631713},
}

@inproceedings{wang2020pm25gnn,
  title     = {{{PM}$_{2.5}$-{GNN}}: A domain knowledge enhanced graph neural network for {PM}$_{2.5}$ forecasting},
  author    = {Wang, Shuo and Li, Yanran and Zhang, Jiang and Meng, Qingye and Meng, Lingwei and Gao, Fei},
  booktitle = {Proceedings of the 28th International Conference on Advances in Geographic Information Systems},
  pages     = {163--166},
  series    = {{SIGSPATIAL} '20},
  year      = {2020},
  doi       = {10.1145/3397536.3422208},
  url       = {https://doi.org/10.1145/3397536.3422208},
}

@inproceedings{hettige2024airphynet,
  title     = {{AirPhyNet}: Harnessing Physics-Guided Neural Networks for Air Quality Prediction},
  author    = {Hettige, Kethmi Hirushini and Ji, Jiahao and Xiang, Shili and Long, Cheng and Cong, Gao and Wang, Jingyuan},
  booktitle = {International Conference on Learning Representations},
  year      = {2024},
  url       = {https://openreview.net/forum?id=JW3jTjaaAB},
}

@inproceedings{wang2021attentive,
  title     = {Modeling inter-station relationships with attentive temporal graph convolutional network for air quality prediction},
  author    = {Wang, Chunyang and Zhu, Yanmin and Zang, Tianzi and Liu, Haobing and Yu, Jiadi},
  booktitle = {Proceedings of the 14th {ACM} International Conference on Web Search and Data Mining},
  pages     = {616--634},
  year      = {2021},
  doi       = {10.1145/3437963.3441731},
}

@article{li2023physics,
  title   = {Improving air quality assessment using physics-inspired deep graph learning},
  author  = {Li, Lianfa and Wang, Jinfeng and Franklin, Meredith and Yin, Qian and Wu, Jiajie and Camps-Valls, Gustau and Zhu, Zhiping and Wang, Chengyi and Ge, Yong and Reichstein, Markus},
  journal = {npj Climate and Atmospheric Science},
  volume  = {6},
  number  = {1},
  pages   = {152},
  year    = {2023},
  doi     = {10.1038/s41612-023-00475-3},
}

@article{pcdcnet2025,
  title   = {{PCDCNet}: A Surrogate Model for Air Quality Forecasting with Physical-Chemical Dynamics and Constraints},
  author  = {Wang, Shuo and Cheng, Yun and Meng, Qingye and Saukh, Olga and Zhang, Jiang and Fan, Jingfang and Zhang, Yuanting and Yuan, Xingyuan and Thiele, Lothar},
  journal = {arXiv preprint arXiv:2505.19842},
  year    = {2025},
  eprint  = {2505.19842},
  archivePrefix = {arXiv},
  primaryClass  = {cs.LG},
  url     = {https://arxiv.org/abs/2505.19842},
}

@article{murray2019bayesian,
  title   = {A {Bayesian} ensemble approach to combine {PM}$_{2.5}$ estimates from statistical models using satellite imagery and numerical model simulation},
  author  = {Murray, Nancy L. and Holmes, Heather A. and Liu, Yang and Chang, Howard H.},
  journal = {Environmental Research},
  volume  = {178},
  pages   = {108601},
  year    = {2019},
  doi     = {10.1016/j.envres.2019.108601},
}

@article{rasp2018neural,
  title   = {Neural networks for postprocessing ensemble weather forecasts},
  author  = {Rasp, Stephan and Lerch, Sebastian},
  journal = {Monthly Weather Review},
  volume  = {146},
  number  = {11},
  pages   = {3885--3900},
  year    = {2018},
  doi     = {10.1175/MWR-D-18-0187.1},
}

@inproceedings{garnelo2018conditional,
  title     = {Conditional neural processes},
  author    = {Garnelo, Marta and Rosenbaum, Dan and Maddison, Christopher and Ramalho, Tiago and Saxton, David and Shanahan, Murray and Teh, Yee Whye and Rezende, Danilo and Eslami, SM Ali},
  booktitle = {Proceedings of the 35th International Conference on Machine Learning},
  series    = {Proceedings of Machine Learning Research},
  volume    = {80},
  pages     = {1704--1713},
  publisher = {{PMLR}},
  year      = {2018},
  url       = {https://proceedings.mlr.press/v80/garnelo18a.html},
}

@inproceedings{gordon2020convolutional,
  title     = {Convolutional conditional neural processes},
  author    = {Gordon, Jonathan and Bruinsma, Wessel P. and Foong, Andrew Y. K. and Requeima, James and Dubois, Yann and Turner, Richard E.},
  booktitle = {International Conference on Learning Representations},
  year      = {2020},
  url       = {https://openreview.net/forum?id=Skey4eBYPS},
}

@article{vaughan2022convcnp,
  title   = {Convolutional conditional neural processes for local climate downscaling},
  author  = {Vaughan, Anna and Tebbutt, Will and Hosking, J. Scott and Turner, Richard E.},
  journal = {Geoscientific Model Development},
  volume  = {15},
  number  = {1},
  pages   = {251--268},
  year    = {2022},
  doi     = {10.5194/gmd-15-251-2022},
}

@article{lam2023graphcast,
  title   = {Learning skillful medium-range global weather forecasting},
  author  = {Lam, Remi and Sanchez-Gonzalez, Alvaro and Willson, Matthew and Wirnsberger, Peter and Fortunato, Meire and Alet, Ferran and Ravuri, Suman and Ewalds, Timo and Eaton-Rosen, Zach and Hu, Weihua and others},
  journal = {Science},
  volume  = {382},
  number  = {6677},
  pages   = {1416--1421},
  year    = {2023},
  doi     = {10.1126/science.adi2336},
}

@article{bi2023pangu,
  title   = {Accurate medium-range global weather forecasting with {3D} neural networks},
  author  = {Bi, Kaifeng and Xie, Lingxi and Zhang, Hengheng and Chen, Xin and Gu, Xiaotao and Tian, Qi},
  journal = {Nature},
  volume  = {619},
  number  = {7970},
  pages   = {533--538},
  year    = {2023},
  doi     = {10.1038/s41586-023-06185-3},
}

@book{cressie2015statistics,
  title     = {Statistics for Spatial Data},
  author    = {Cressie, Noel},
  publisher = {John Wiley \& Sons},
  year      = {2015},
}

@inproceedings{kim2019attentive,
  title     = {Attentive neural processes},
  author    = {Kim, Hyunjik and Mnih, Andriy and Schwarz, Jonathan and Garnelo, Marta and Eslami, Ali and Rosenbaum, Dan and Vinyals, Oriol and Teh, Yee Whye},
  booktitle = {International Conference on Learning Representations},
  year      = {2019},
  url       = {https://openreview.net/forum?id=SkE6PjC9KX},
}

@inproceedings{ronneberger2015unet,
  title     = {{U-Net}: Convolutional Networks for Biomedical Image Segmentation},
  author    = {Ronneberger, Olaf and Fischer, Philipp and Brox, Thomas},
  booktitle = {International Conference on Medical Image Computing and Computer-Assisted Intervention},
  series    = {Lecture Notes in Computer Science},
  volume    = {9351},
  pages     = {234--241},
  year      = {2015},
  doi       = {10.1007/978-3-319-24574-4_28},
}

@inproceedings{vaswani2017attention,
  title     = {Attention is all you need},
  author    = {Vaswani, Ashish and Shazeer, Noam and Parmar, Niki and Uszkoreit, Jakob and Jones, Llion and Gomez, Aidan N and Kaiser, {\L}ukasz and Polosukhin, Illia},
  booktitle = {Advances in Neural Information Processing Systems},
  volume    = {30},
  pages     = {5998--6008},
  year      = {2017},
  url       = {https://proceedings.neurips.cc/paper/2017/hash/3f5ee243547dee91fbd053c1c4a845aa-Abstract.html},
}

@article{wang2004ssim,
  title   = {Image quality assessment: from error visibility to structural similarity},
  author  = {Wang, Zhou and Bovik, Alan C. and Sheikh, Hamid R. and Simoncelli, Eero P.},
  journal = {{IEEE} Transactions on Image Processing},
  volume  = {13},
  number  = {4},
  pages   = {600--612},
  year    = {2004},
  doi     = {10.1109/TIP.2003.819861},
}

@article{farr2007srtm,
  title   = {The {Shuttle Radar Topography Mission}},
  author  = {Farr, Tom G. and Rosen, Paul A. and Caro, Edward and Crippen, Robert and Duren, Riley and Hensley, Scott and Kobrick, Michael and Paller, Mimi and Rodriguez, Ernesto and Roth, Ladislav and others},
  journal = {Reviews of Geophysics},
  volume  = {45},
  number  = {2},
  pages   = {RG2004},
  year    = {2007},
  doi     = {10.1029/2005RG000183},
}

@inproceedings{loshchilov2017decoupled,
  title     = {Decoupled weight decay regularization},
  author    = {Loshchilov, Ilya and Hutter, Frank},
  booktitle = {International Conference on Learning Representations},
  year      = {2019},
  url       = {https://openreview.net/forum?id=Bkg6RiCqY7},
}

@incollection{shapley1953value,
  title     = {A value for {n}-person games},
  author    = {Shapley, Lloyd S.},
  booktitle = {Contributions to the Theory of Games II},
  editor    = {Kuhn, Harold W. and Tucker, Albert W.},
  publisher = {Princeton University Press},
  address   = {Princeton, NJ},
  pages     = {307--317},
  year      = {1953},
}

@inproceedings{lundberg2017unified,
  title     = {A unified approach to interpreting model predictions},
  author    = {Lundberg, Scott M and Lee, Su-In},
  booktitle = {Advances in Neural Information Processing Systems},
  volume    = {30},
  pages     = {4765--4774},
  year      = {2017},
  url       = {https://proceedings.neurips.cc/paper/2017/hash/8a20a8621978632d76c43dfd28b67767-Abstract.html},
}

@techreport{horalek2006spatial,
  title       = {Spatial mapping of air quality for {European} scale assessment},
  author      = {Hor{\'a}lek, Jan and Denby, Bruce and de Smet, Peter and de Leeuw, Frank A. A. M. and Kurf{\"u}rst, Pavel and Swart, Rob and van Noije, Twan},
  institution = {European Topic Centre on Air and Climate Change},
  number      = {{ETC/ACC} Technical Paper 2006/6},
  address     = {Bilthoven, The Netherlands},
  year        = {2007},
  url         = {https://www.eionet.europa.eu/etcs/etc-atni/products/etc-atni-reports/etcacc_technpaper_2006_6_spat_aq},
}

@inproceedings{le2020spatiotemporal,
  title        = {Spatiotemporal deep learning model for citywide air pollution interpolation and prediction},
  author       = {Le, Van-Duc and Bui, Tien-Cuong and Cha, Sang-Kyun},
  booktitle    = {2020 {IEEE} International Conference on Big Data and Smart Computing ({BigComp})},
  pages        = {55--62},
  year         = {2020},
  organization = {{IEEE}},
  doi          = {10.1109/BigComp48618.2020.00-99},
}

@inproceedings{zheng2015forecasting,
  title     = {Forecasting fine-grained air quality based on big data},
  author    = {Zheng, Yu and Yi, Xiuwen and Li, Ming and Li, Ruiyuan and Shan, Zhangqing and Chang, Eric and Li, Tianrui},
  booktitle = {Proceedings of the 21st {ACM SIGKDD} International Conference on Knowledge Discovery and Data Mining},
  pages     = {2267--2276},
  year      = {2015},
  doi       = {10.1145/2783258.2788573},
}

@inproceedings{scholz2023sim2real,
  title     = {{Sim2Real} for environmental neural processes},
  author    = {Scholz, Jonas and Andersson, Tom R and Vaughan, Anna and Requeima, James and Turner, Richard E},
  booktitle = {NeurIPS 2023 Workshop on Tackling Climate Change with Machine Learning},
  year      = {2023},
  url       = {https://www.climatechange.ai/papers/neurips2023/77},
}

@article{allen2025end,
  title   = {End-to-end data-driven weather prediction},
  author  = {Allen, Anna and Markou, Stratis and Tebbutt, Will and Requeima, James and Bruinsma, Wessel P. and Andersson, Tom R. and Herzog, Michael and Lane, Nicholas D. and Chantry, Matthew and Hosking, J. Scott and others},
  journal = {Nature},
  volume  = {641},
  number  = {8065},
  pages   = {1172--1179},
  year    = {2025},
  doi     = {10.1038/s41586-025-08897-0},
}

\end{document}